\documentclass[12pt,conference]{IEEEtran}
\usepackage[latin9]{inputenc}
\usepackage{amsmath}
\usepackage{amssymb}
\usepackage{graphicx}

\makeatletter

\providecommand{\tabularnewline}{\\}

\makeatother

\begin{document}
\title{Prognostics and Health Management of Wafer Chemical-Mechanical Polishing
System using Autoencoder}
\author{\IEEEauthorblockN{Kart-Leong Lim} \IEEEauthorblockA{\textit{Institute of Microelectronics} \\
 A{*}Star, Singapore \\
 Email: limkl@ime.a-star.edu.sg} \and \IEEEauthorblockN{Rahul Dutta} \IEEEauthorblockA{\textit{Institute of Microelectronics} \\
 A{*}Star, Singapore \\
 Email: dutta@ime.a-star.edu.sg}}
\maketitle
\begin{abstract}
The Prognostics and Health Management Data Challenge (PHM) 2016 tracks
the health state of components of a semiconductor wafer polishing
process. The ultimate goal is to develop an ability to predict the
measurement on the wafer surface wear through monitoring the components
health state. This translates to cost saving in large scale production.
The PHM dataset contains many time series measurements not utilized
by traditional physics based approach. On the other hand task, applying
a data driven approach such as deep learning to the PHM dataset is
non-trivial. The main issue with supervised deep learning is that
class label is not available to the PHM dataset. Second, the feature
space trained by an unsupervised deep learner is not specifically
targeted at the predictive ability or regression. In this work, we
propose using the autoencoder based clustering whereby the feature
space trained is found to be more suitable for performing regression.
This is due to having a more compact distribution of samples respective
to their nearest cluster means. We justify our claims by comparing
the performance of our proposed method on the PHM dataset with several
baselines such as the autoencoder as well as state-of-the-art approaches. 
\end{abstract}

\section{Introduction}

\begin{figure*}
\begin{centering}
\includegraphics[scale=0.31]{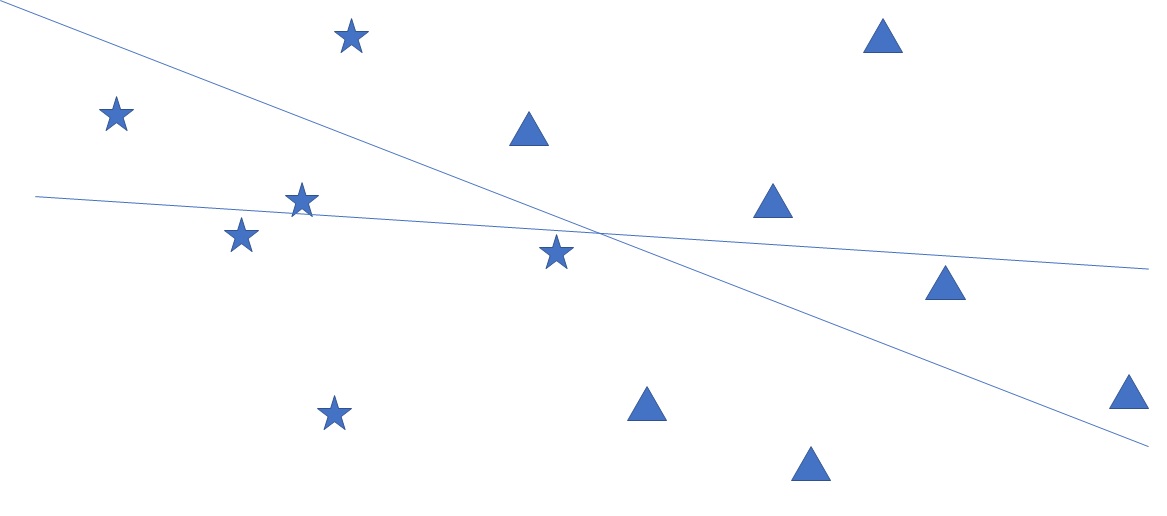}$\;$$\;$$\;$$\;$$\;$$\;$$\;$$\;$$\;$$\;$$\;$$\;$\includegraphics[scale=0.4]{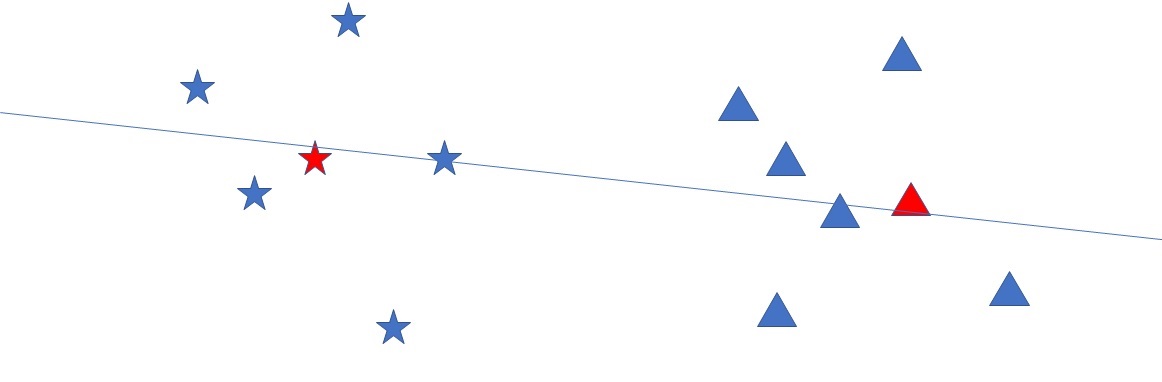} 
\par\end{centering}
\caption{\textbf{Left}: Reconstruction loss based autoencoder latent space
is not optimized for regression. \textbf{Right}: Clustering in the
latent space can improve regression due to the samples being more
sparsely distributed (with respect to their cluster mean, as depicted
in red).}
\end{figure*}

The semiconductor industry is a multi-billions industry which manufactures
nanoscale processors that is essential for modern connectivity and
productivity. Large semiconductor companies such as Intel and TSMC
manufacture microchips for the world's smart devices and automobiles.
The Prognostics and Health Management Data Challenge (PHM) 2016\footnote{https://www.phmsociety.org/events/conference/phm/16/data-challenge}
tracks the health state of components of a chemical mechanical polishing
(CMP) process. The ultimate goal is to develop an ability to predict
the measurement on the surface wear a.k.a material removal rate (MRR)
through monitoring the components health state. This ability translates
to large scale cost estimates and preplanning such as raw supply estimates,
delivery schedules, overhead cost estimates and etc. The semiconductor
industry seeks to improve its current automated processes through
using state-of-the-arts techniques in deep learning. One such process
is the CMP where the prediction of material removal rate in wafer
is desired. The forefront of deep learning research is often found
in the domain of visual recognition. The success of deep learning
approaches such as ResNet \cite{he2016deep}, VAE \cite{kingma2014stochastic},
GAN \cite{goodfellow2014generative} in recent years in computer vision
has captured the attention of many researchers from other domains.
The next logical step in deep learning would be to see its deployment
into the large funding sectors such as the semiconductor, automobile,
military, health, finance and etc. In particular, the fabrication
of wafers involves over several hundreds of processes. Each process
is complex and costly to operate and not all processes are automated.
Human intervention often results in productivity loss. In CMP, a skilled
operator is required to periodically halt the machine in order to
manually track the depthness of polished wafer surface, or the MRR.
From an artificial intelligence standpoint, the CMP process contains
many components health state not fully exploited by the human but
exploitable by deep learning. These components health state can be
grouped under the families of usage, pressure, slurry and rotation
as seen in Fig 4. As the concatenated raw dimension in total is very
large at over several thousands, many prior works turn their focus
on feature selection or feature extraction, prior to regression.

Regression in CMP models the relationship between the continuous variable
MRR prediction and the CMP measurement in time series. When we have
an original input space or high dimensional input, standard regression
model alone may not effectively capture the relationship between both
input and output. Instead of pursuing the direction of more sophisticated
statistical methods for regression, another common approach is to
perform feature selection/ dimension reduction prior to regression.
More specifically, we consider using deep learning for feature selection.
However, there are two issues: 
\begin{verse}
i) Regression dataset cannot be trained easily by supervised deep
learning.

ii) Reconstruction loss alone may not be meaningful for regression. 
\end{verse}
First, regression do not rely on class labels for training. Thus,
it is difficult to apply traditional deep learning such as CNN to
regression datasets. Most regression dataset do not have class labels,
thus the task is to more suitable for unsupervised type of deep learning
such as the autoencoder. Second, the feature space trained by an autoencoder
using reconstruction loss alone may not be suitable for regression
because it is tasked with self correction based learning instead of
specifically addressing the regression task. Typically, reconstruction
loss will cause the distribution of samples in the latent space to
overlap such as in MNIST \cite{hinton2006reducing}. In the worst
scenario, we may see several suboptimal linear regression lines when
performing regression in the latent space as seen in Fig 1. Thus,
we require exploring other efficient loss function for the autoencoder.
Specifically, we use an autoencoder loss function known as the autoencoder
based clustering (ABC) to train the autoencoder \cite{song2013auto}.
This approach focuses on minimizing the difference between the distribution
of the samples by the encoder and the partitioning by Kmeans in the
autoencoder latent space. As illustrated in Fig 1, when we have a
latent space that is optimized using class distribution, linear regression
should improve since it is easier to find an optimal line due to an
overall more compact representation. In a practical scenario, we may
not have ground truth on how many clusters to use in the latent space.
An extension is to consider the infinite Gaussian mixture model (iGMM),
which simultaneously solve both model selection and clustering for
ABC's deep clustering.

We demonstrated our proposed deep learning approach mainly on the
PHM Challenge 2016. In the PHM dataset, our proposed approach is used
as a feature selection taking the variables from the CMP process as
raw inputs, before training a linear regression in the feature or
latent space. We compare our proposed approach with two different
baseline methods using raw inputs directly and statistical moments
from the raw inputs. From experimental result, it shows that our deep
learning approach outperforms both baselines. When compared to the
state-of-the-arts on the PHM dataset, our method is better than methods
using features from statistical moments, Random Forest to select samples
from training, closed-form physics-based model as well as other non
deep learning based strategies.

\section{Methodology}

\subsection{Autoencoder based clustering loss}

Kmeans alone only computes cluster mean and cluster assignment in
the latent space and have zero influence over learning the weight,
$w$ of the encoder. Similarly for autoencoder (AE), there is no way
the encoder trained using reconstruction loss or mean square error
(MSE) alone can approach a Kmeans partitioning. The autoencoder based
clustering (ABC) \cite{song2013auto} approach unites both by introducing
a clustering loss. The learning in ABC is obtained through (shown
in Fig 2) 
\begin{verse}
i) reconstruction loss - standard mean square error on the target
which is the encoder input vs the network output from the decoder.

ii) clustering loss - minimizing the error between the nearest cluster
mean $B^{*}$ (nearest is denoted by {*}) and the encoder output $z$,
as a point in the latent space (input $x$ that passes through the
encoder). 
\end{verse}
Conceptually, the distribution of the samples in the latent space
will become closer to their nearest cluster means as illustrated in
Fig 1.

Formally, the cluster mean, sample in the latent space and sample
$x$ in the original dimension $D$ are denoted by $B=\left\{ B_{k}\right\} _{k=1}^{K}\in\mathbb{R}{}^{Z}$,
$z=\left\{ z_{n}\right\} _{n=1}^{N}\in\mathbb{R}^{Z}$ and $x=\left\{ x_{n}\right\} _{n=1}^{N}\in\mathbb{R}^{D}$
respectively. $x$ refers to target/ input and $y$ is the network
output. The encoder output, $z$ is computed using the last row of
eqn (1). We can replace $x$ with $z$ for computing the decoder.
$f_{1}$ and $f_{2}$ are the $tanh$ and $sigmoid$ activation functions
respectively. We define the ABC loss function in \cite{song2013auto,lim2020}
as follows

\begin{equation}
\begin{array}{c}
\mathcal{L}{}_{ABC}=Reconst.\;loss+clustering\;loss\\
=-\frac{1}{2}\left(x-y\right)^{2}+\frac{1}{2}\left(z-B^{*}\right)^{2}\\
\\
z=f_{2}\left(\sum_{h+1}w_{h+1,h+2}\cdot f_{1}\left(\sum_{h}w_{h,h+1}\cdot x_{h}\right)\right)\\
\\
\end{array}
\end{equation}
\\

We train eqn (1) using SGD with momentum where $w_{h,h+1}$ refers
to the weight between layer $h$ and $h+1$ below

\begin{equation}
\begin{array}{c}
\left(w_{h,h+1}\right)_{t}=\left(w_{h,h+1}\right)_{t-1}-\eta\left(V_{h,h+1}\right)_{t}\\
\\
\left(V_{h,h+1}\right)_{t}=\rho\left(V_{h,h+1}\right)_{t-1}+(1-\rho)*\frac{\delta\mathcal{L}{}_{ABC}}{\delta w_{h,h+1}}
\end{array}
\end{equation}

\subsection{Clustering approach}

When we model a dataset (regardless of original space or latent space)
using unsupervised learning such as Kmeans, we can represent the dataset
as a set of hidden variables. These variables can in turn be used
to train other task e.g. Bag-of-Words feature extraction \cite{csurka2004visual},
SVM-KNN classifier \cite{zhang2006svm} or DP-mean clustering \cite{Kulis2012}
and etc. More specifically in this paper, the cluster mean is fed
into $\mathcal{L}{}_{ABC}$ to train the AE.

\begin{figure*}
\begin{centering}
\includegraphics[scale=0.56]{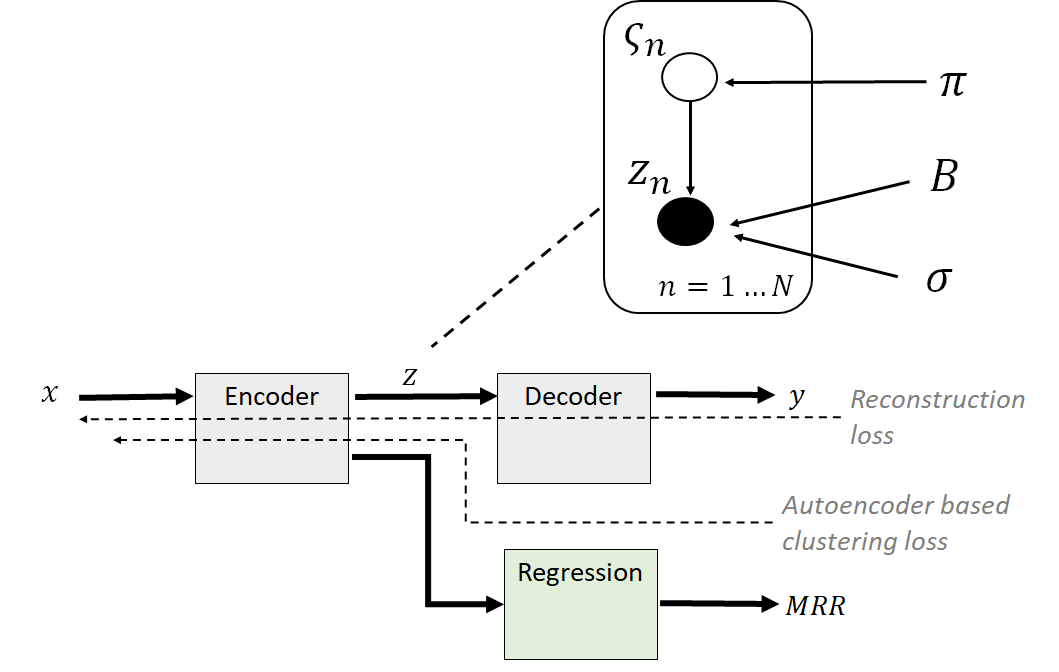} 
\par\end{centering}
\caption{Backpropagation training (dashed arrows) of the raw latent space $z$
using reconstruction loss and ABC loss. The ABC loss exploits GMM
hidden variables $B,\sigma$ and $\pi$ for the encoder weight training.
Finally, MRR regression modeling and prediction are performed in the
latent space of the trained encoder.}
\end{figure*}

\subsubsection{Kmeans }

Our main clustering algorithm is the Kmeans algorithm where cluster
size $K$ is assumed given. The cluster mean $B$ and cluster assignment
$\varsigma$ are point estimated as below

\begin{equation}
\begin{array}{c}
\hat{B_{k}}=\frac{\sum_{n=1}^{N}\hat{\varsigma_{nk}}z_{n}}{\sum_{n=1}^{N}\hat{\varsigma_{nk}}}\\
\\
\hat{\varsigma_{nk}}=\underset{k}{\arg\max}\;\left\{ -\frac{1}{2}\left(z_{n}-\hat{B_{k}}\right){}^{2}\right\} \varsigma_{nk}
\end{array}
\end{equation}
We denoted $B=\left\{ B_{k}\right\} _{k=1}^{K}\in\mathbb{R}{}^{Z}$
and $\varsigma=\left\{ \varsigma_{n}\right\} _{n=1}^{N}$ where $\sum_{k=1}^{K}\varsigma_{nk}=1$
and $\varsigma_{nk}\in\left\{ 0,1\right\} $. \\

\subsubsection{Infinite Gaussian mixture model}

There are two key advantages over Kmeans in clustering: 
\begin{verse}
i) The number of clusters can be automatically found.

ii) Each cluster can vary its variance. 
\end{verse}
We can define the standard mixture of Gaussian as an infinite mixture
of Gaussians as follows

\begin{equation}
\ln p(z|B,\sigma,\pi)=\sum_{n=1}^{N}\sum_{k=1}^{T=\infty}\left\{ \ln\pi_{k}+\ln\mathcal{N}(z_{n}|B_{k},\sigma_{k})\right\} \varsigma_{nk}
\end{equation}
\\

We denote cluster mixture component as $\pi=\left\{ \pi_{k}\right\} _{k=1}^{T=\infty}\in\mathbb{R}{}^{Z}$,
cluster mean $B=\left\{ B_{k}\right\} _{k=1}^{T=\infty}\in\mathbb{R}{}^{Z}$
and cluster variance as $\sigma=\left\{ \sigma_{k}\right\} _{k=1}^{T=\infty}\in\mathbb{R}{}^{Z}$.

The expectation-maximization algorithm of eqn (4) allows us to obtain
a set of closed form equations \cite{bishop2006pattern,Lim2016}

\begin{equation}
\begin{array}{c}
\hat{B_{k}}=\frac{\sum_{n=1}^{N}\hat{\varsigma_{nk}}z_{n}}{\sum_{n=1}^{N}\hat{\varsigma_{nk}}}\\
\\
\hat{\varsigma_{nk}}=\underset{k}{\arg\max}\;\left\{ \ln\pi_{k}-\frac{1}{2}\left(z_{n}-\hat{B_{k}}\right){}^{2}\right\} \varsigma_{nk}\\
\\
\hat{\pi_{k}}=\frac{\sum_{n=1}^{N}\varsigma_{nk}}{\sum_{k=1}^{K}\sum_{n=1}^{N}\varsigma_{nk}}\\
\\
\end{array}
\end{equation}
\\
In the standard GMM, the significance of $\pi$ is that each cluster
is represented as a probability $\pi_{k}$ as opposed to hard assignment
in Kmeans.

Another important use of $\pi$ is cluster pruning in \cite{Lim2016}.
In practice, for infinite GMM we cannot possibly work with an infinite
number of clusters. Instead, we set $T$ to be a sufficiently large
value. Then, each iteration of eqn (5), we discard the $k^{th}$ cluster
that contains insignificant values in $\hat{\pi_{k}}$ and update
$T$. At optimality, $T$ approaches ground truth $K$.

Lastly as $\mathcal{L}{}_{ABC}$ is unable to utilize the $\sigma$
term in GMM, we simply keep it constant for each GMM cluster (a.k.a
shared diagonal covariance GMM) \cite{Lim2016}.

\section{Experiment: Deep clustering for predictive modeling}

The PHM Challenge 2016 dataset which collect measurements from a CMP
process contains in total 2556 samples in Table I with mainly 18 time
series variables. The evaluation of our regression model prediction
is using root mean square error (RMSE) \cite{yu2019predictive}.

\begin{table}
\begin{centering}
\begin{tabular}{|c|c|c|}
\cline{2-3} \cline{3-3} 
\multicolumn{1}{c|}{} & \multicolumn{2}{c|}{RMSE}\tabularnewline
\cline{2-3} \cline{3-3} 
\multicolumn{1}{c|}{} & Low Wear  & High Wear\tabularnewline
\hline 
Train  & 1616  & 166\tabularnewline
Valid.  & 354  & 34\tabularnewline
Test  & 354  & 32\tabularnewline
\hline 
\end{tabular}
\par\end{centering}
\caption{PHM 2016 Challenge}
\end{table}

\subsection{Feature extraction}

\subsubsection{Statistical moments on time series}

We refer to Moment18x4 as our baseline approach using the first 4
central moments for each 18 variables, i.e mean, standard deviation,
skewness and kurtosis on the PHM dataset in Fig 3. Thus, the extracted
dimension is at 72.

Subsequently, we removed 6 time series variables retaining only 12
variables below, for reasons which we will discuss shortly. We then
apply the first 4 central moments to these 12 which we refer to as
Moment12x4 with a dimension at 48. 
\begin{quote}
Usages: backing film, dresser, dresser table, membrane, pressurized
sheet, polishing table.

Pressures: pressurized chamber, main outer air bag, center air bag,
retainer ring, ripple air bag and edge air bag. 
\end{quote}

\subsubsection{Concatenated time series }

In \cite{9315140} the authors used a feature extraction approach,
FE2 which concatenates the edge airbag pressure and retainer ring
pressure. Intuitively, this approach measures the physical contact
between the table that holds the wafer (edge airbag pressure) and
the wafer carrier which exert force on the wafer (retainer ring pressure).
Subsequently, they extended FE2 to FE12 using the 12 variables described
above. In our third baseline approach, we directly apply the raw input
of FE12 to the AE. Briefly explaining FE12, it is constructed as follows:
Starting from a time series of 400 dimensions per variable, we downsampled
each time series by a factor of 8, to a dimension of 50, which returns
a concatenated vector of 600 dimensions. We use downsampling on the
time series data due to high dimensions. As a result we may tradeoff
faster computation for slight loss of accuracy in MRR prediction.
Lastly, we normalization the vector. We rename the FE12 appproach
in \cite{9315140} as RawSpace12x50 in this paper.

\subsection{Autoencoder setup}

Our architecture is using $600-500-100$ for the encoder and vice
versa for the decoder, taking RawSpace12x50 as raw input which has
a RMSE at 7.9504 for low wear in Table II. The number of iterations
for the clustering and reconstruction losses are capped at 1000 each.
For regression, we find that using linear regression in the latent
space works best for low wear RMSE. We used stochastic gradient ascent
(SGA) with momentum (where $\rho$=0.98) as it outperforms plain vanilla
SGA for MRR prediction. We use a minibatch of 2 samples for the weights
training since the dataset is quite small. Our Kmeans minibatch size
is 40 samples. Empirically, we found that selecting $K=2$ clusters
works best for the PHM dataset. Lastly for high wear, we reused the
network weights we obtained for low wear.

\subsection{Baselines results}

In Table II, we trained a linear regression model on the training
set (we combined training and validation set) for low wear and high
wear separately on Moment18x4, Moment12x4 and RawSpace12x50. The averaged
result (over 10 attempts) on the test dataset shows that Moment12x4
outperforms Moment18x4 in terms of root mean square error (RMSE).
We suspect that slurry and rotation families add redundancy/ noise
to the regression model and actually worsen the RMSE. Overall, the
RMSE is the best on RawSpace12x50. This also shows that solely using
central moments is not ideal as there are loss of representation unlike
using the raw dimension, with the tradeoff being a larger dimension.
In our next baseline attempt, PCA30, we use prinicipal component analysis
on RawSpace12x50. Empirically, we found that retaining the top 30
largest eigenvalues works best. Overall, PCA30 has lower RMSE than
RawSpace12x50.

\begin{table}
\begin{centering}
\begin{tabular}{|c|c|c|}
\cline{2-3} \cline{3-3} 
\multicolumn{1}{c|}{} & \multicolumn{2}{c|}{RMSE}\tabularnewline
\cline{2-3} \cline{3-3} 
\multicolumn{1}{c|}{} & Low Wear  & High Wear\tabularnewline
\hline 
Baseline \#1: Moment18x4  & 12.8723  & -\tabularnewline
Baseline \#2: Moment12x4  & 8.4934  & -\tabularnewline
Baseline \#3: RawSpace12x50  & 7.9504  & 5.3205\tabularnewline
Baseline \#4: PCA30  & 8.1794  & 3.9687\tabularnewline
\hline 
\end{tabular}
\par\end{centering}
\caption{Baseline methods}
\end{table}

\subsection{Reconstruction loss vs Clustering loss}

Reconstruction loss alone may not be meaningful for regression because
it is tasked with self correction based learning instead of specifically
addressing the regression task. We setup two different AEs to have
identical random initial weight. We then record the RMSE value when
a regression model is trained in the latent space of each AE using
identical random initial weight. Then we trained separately, i) AE
with reconstruction loss and ii) AE with clustering loss using $K=2$,
both on the low wear training plus validation dataset in Table 1.
In Fig 1 and Fig 3, the main idea is to bin the entire MRR range so
that when performing clustering on each bin, the samples within each
cluster should ideally be as close to the cluster center as possible.
As a consequence, we minimize the within distance of each cluster,
allowing regression modeling and prediction to improve.

In Table III, we tabulate the results of regression in the latent
space of i) and ii) using the test set. We observed that while reconstruction
loss can reduce the RMSE on attempt \#2 and \#3, the improvement is
not as significant as when compared to clustering loss.

\begin{table}
\begin{centering}
\begin{tabular}{|c|c|c|c|}
\cline{2-4} \cline{3-4} \cline{4-4} 
\multicolumn{1}{c|}{} & \multicolumn{3}{c|}{RMSE, Low Wear}\tabularnewline
\cline{2-4} \cline{3-4} \cline{4-4} 
\multicolumn{1}{c|}{} & \#1  & \#2  & \#3\tabularnewline
\hline 
Initial random weights  & 6.5568  & 7.1297  & 6.4517\tabularnewline
Reconst. loss  & 6.5934  & 6.9606  & 6.2542\tabularnewline
Clust. loss (Kmean)  & \textbf{5.9704}  & \textbf{6.5979}  & \textbf{5.9704}\tabularnewline
\hline 
\end{tabular}
\par\end{centering}
\caption{Reconstruction loss vs clustering loss}
\end{table}

\subsection{Clustering loss: Infinite GMM vs Kmeans}

When we consider the MRR histogram in Fig 3, it is not intuitive to
know how many $K$ clusters we should ideally use in the clustering
loss in eqn (1), as the sample size is very low at each end of the
distribution for low wear MRR. When we use too many clusters, there
will be insufficient statistics for clusters with low sample count.

For the clustering loss using Kmeans approach, we ran each fixed cluster
sizes from $K$= 2 to 10. We manually found the best RMSE is at $K=2$.
We call this value as ground truth (gt) in Table IV.

On the other hand, we employ the infinite GMM clustering for the clustering
loss. We initially set $T=10$. After optimization of both weight
and infinite GMM via cluster pruning, we obtained the estimated clustering
size to be between 3 and 4. In Table IV, we showed that the RMSE of
the clustering loss using the infinite GMM outperforms the clustering
loss using Kmeans. This is because infinite GMM uses soft assignment
and infinite mixture as compared to hard assignment in Kmeans. A disadvantage
of using this method is the additional computational overhead over
the Kmeans counterpart. Furthermore on the PHM dataset, due to the
comparatively small training size, the performance gained in lower
RMSE is not significant.

\begin{figure*}
\begin{centering}
\includegraphics[scale=0.65]{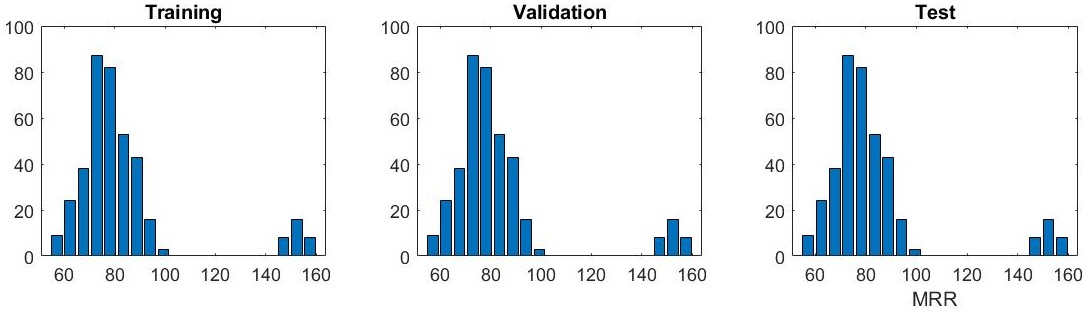} 
\par\end{centering}
\caption{The MRR histograms of wafers. We observed two distributions which
refer to samples with low wear ($50\protect\leq MRR\protect\leq100$)
and high wear ($140\protect\leq MRR\protect\leq200$) respectively
for each dataset partitioning (i.e. training, validation and test)}
\end{figure*}

\begin{table}
\begin{centering}
\begin{tabular}{|c|cc|cc|cc|}
\cline{2-7} \cline{3-7} \cline{4-7} \cline{5-7} \cline{6-7} \cline{7-7} 
\multicolumn{1}{c|}{} & \multicolumn{6}{c|}{RMSE, Low Wear}\tabularnewline
\cline{2-7} \cline{3-7} \cline{4-7} \cline{5-7} \cline{6-7} \cline{7-7} 
\multicolumn{1}{c|}{} & \multicolumn{2}{c|}{\#1} & \multicolumn{2}{c|}{\#2} & \multicolumn{2}{c|}{\#3}\tabularnewline
\cline{2-7} \cline{3-7} \cline{4-7} \cline{5-7} \cline{6-7} \cline{7-7} 
\multicolumn{1}{c|}{} & ACC  & $\hat{K}$  & ACC  & $\hat{K}$  & ACC  & $\hat{K}$\tabularnewline
\hline 
Ini. random weights  & 6.5568  & -  & 7.1297  & -  & 6.4517  & -\tabularnewline
CL. (Kmean) & 5.9704  & gt  & 6.5979  & gt  & 5.9704  & gt\tabularnewline
CL. ($\infty$GMM) & \textbf{5.8784}  & 3  & \textbf{6.5424}  & 4  & \textbf{5.9590}  & 4\tabularnewline
\hline 
\end{tabular}
\par\end{centering}
\caption{Infinite GMM vs Kmeans (CL refers to clustering loss)}
\end{table}

\subsection{Reconstruction loss + Clustering loss with Kmeans}

We seek a latent space where training a regression model for prediction
can improve the RMSE result. Our proposed approach is using an AE
trained with both losses (shown in Fig 2). As mentioned earlier the
proposed latent space which is trained using class information is
more suitable for regression due to having a more compact distribution
of samples respective to their nearest classes (shown in Fig 1). 

In Table V, we trained the AE separately first with reconstruction
loss. For initial random weights before using reconstruction loss,
in the latent space the linear regression trained model obtained a
RMSE for low wear at 6.8959. Reconstruction loss gradually reduces
to 6.255 at the end of training. Training the latent space with clustering
loss further reduces the RMSE to 5.6872. In the same latent space,
we trained another regression model for high wear and obtained a RMSE
of 3.859 on the test set. The proposed approach outperforms all baselines
from \#1-4 in Table 2 as well as the AEs with individual losses in
Table 3.

\begin{table}
\begin{centering}
\begin{tabular}{|c|c|c|}
\cline{2-3} \cline{3-3} 
\multicolumn{1}{c|}{} & \multicolumn{2}{c|}{RMSE}\tabularnewline
\cline{2-3} \cline{3-3} 
\multicolumn{1}{c|}{} & Low Wear & High Wear\tabularnewline
\hline 
Initial random weights & 6.8959 & -\tabularnewline
Reconst. loss & 6.255 & -\tabularnewline
Reconst. + Clust. loss (Kmean) & \textbf{5.6872} & \textbf{3.8590}\tabularnewline
\hline 
\end{tabular}
\par\end{centering}
\caption{Proposed method}
\end{table}

\subsection{Comparison with state-of-the-arts}

We compare our method with some state-of-the-arts CPM methods in Table
VI. Our best attempt with the proposed loss function achieved an accuracy
of $RMSE=4.77$ for regression prediction (averaged over 10 attempts),
using the approach described in Table 5. Most works \cite{li2019prediction,jia2018adaptive,di2017enhanced,li2018assessment}
extract statistical moments from the time series variables similarly
to our Moment18x4. In addition some authors use features using Fourier
transform. Then, they applied the feature to a base learner such as
linear regression or decision tree. In \cite{li2019prediction} the
authors applied the above mentioned features to a stacking of base
learners trained by Extreme Learning Machine (ELM-stacking). Contrary
to using statistical moments, the approach in \cite{yu2019predictive}
use Random Forest to select samples from training. The samples are
then applied to closed-form physics-based model to perform MRR regression.
Other strategies for feature selection or extraction include using
K nearest neighbor for the usage families \cite{di2017enhanced,jia2018adaptive}
and area under curve of variables \cite{di2017enhanced}, deep belief
net (DBN) \cite{wang2017deep} and self-organizing machine variant
(GMDH) \cite{jia2018adaptive}.

We did not include the results of DBN and GMDH. We first discuss the
result of \cite{wang2017deep} with a RMSE of 2.7 achieved using DBN
for feature extraction and MLP for regression. They also reported
that when using DBN for feature extraction and support vector regression
they could achieve RMSE at 3.1. Although the authors in \cite{wang2017deep}
claimed that only the variables under the usage family is critical
to MRR i.e. excluding all other variables from the pressure, rotation
and slurry families. However when replicating this finding, we could
not reproduce this claim. Instead, we found that using both the usage
and pressure families gives better result than usage family alone
for our AE approach. In the approach of GMDH \cite{jia2018adaptive},
the authors reported that when using linear regression with statistical
moments, they reported a RMSE of 4.03. However, this is way lower
than the RMSE of 12.87 we obtained for Moment18x4. We suspect the
discrepancy between our method and \cite{jia2018adaptive} is likely
due to the finer grouping of the PHM 2016 dataset into 3 smaller groups
using additional information such as chamber ID. We did not pursue
such fine grouping as it may not be realistic under practical scenario.

\begin{table}
\begin{centering}
\begin{tabular}{|c|c|}
\cline{2-2} 
\multicolumn{1}{c|}{} & RMSE\tabularnewline
\multicolumn{1}{c|}{} & Unknown Wear\tabularnewline
\hline 
Preston model \cite{wang2017deep}  & 29.5\tabularnewline
Random Forest \cite{yu2019predictive}  & 16.973\tabularnewline
GBT \cite{li2019prediction}  & 8.252\tabularnewline
ELM-stacking \cite{li2019prediction}  & 7.261\tabularnewline
2nd Physical model \cite{wang2017deep}  & 7.6\tabularnewline
Reconst. + Clust. loss (Kmeans)  & \textbf{4.7736}\tabularnewline
\hline 
\end{tabular}
\par\end{centering}
\caption{Prediction error on PHM 2016}
\end{table}

\begin{figure*}
\begin{centering}
\includegraphics[scale=0.45]{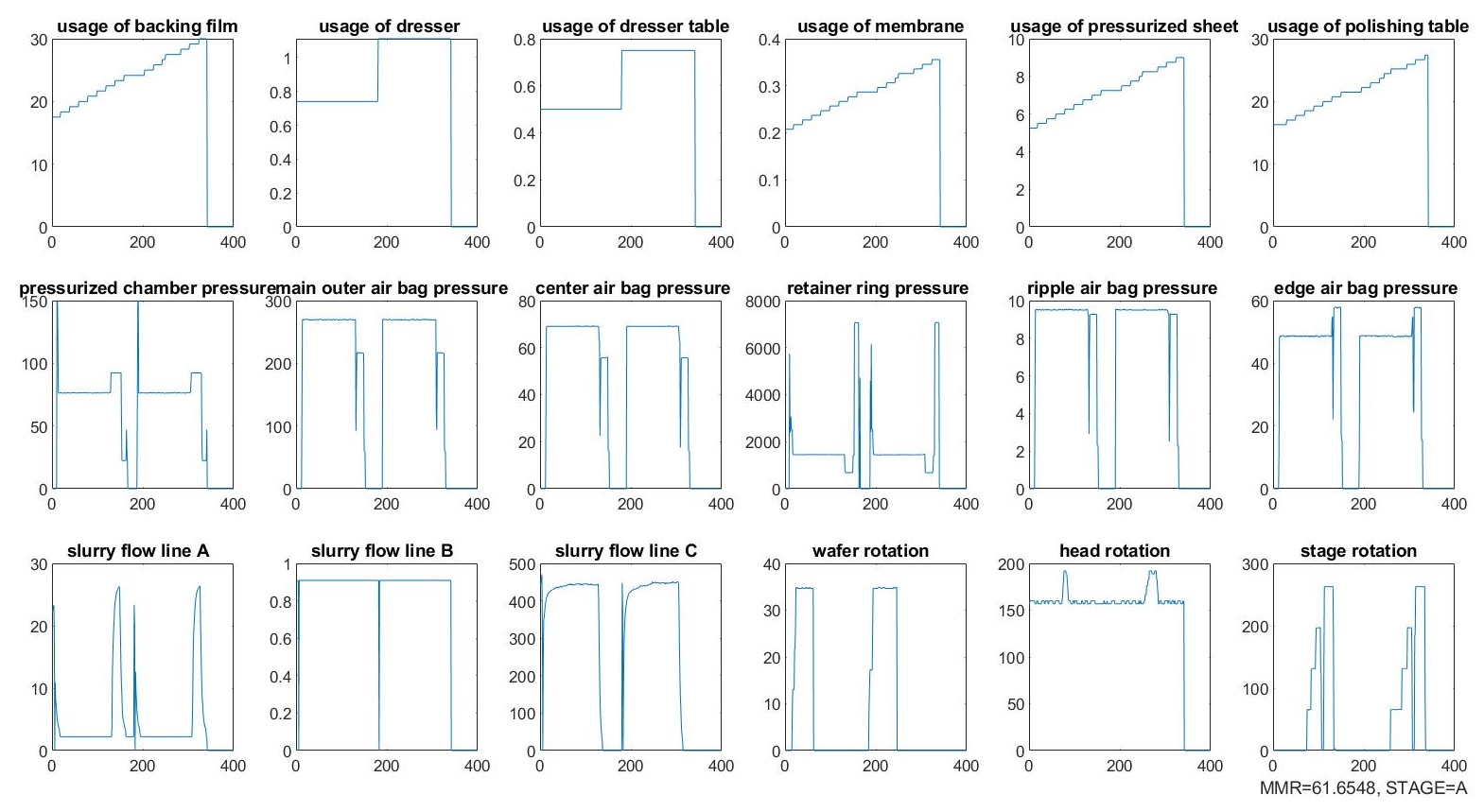}
\par\end{centering}
\caption{Variables of usages, pressure, slurry and rotation of a CMP process
for a wafer sample.}
\end{figure*}

\section{Conclusion}

Feature selection is postulated as the most crucial step in the prediction
of CMP process, whereby the feature space seeks to reduce the error
of MRR prediction. Previous feature selection approaches rely on techniques
such as statistical moments, integration under the curve, decision
tree, PCA, closed-form physics-based model and etc. More recent approaches
use deep learning such as deep belief net and self-organizing machines.
The main problem with using deep learning such as CNN for CMP process
is that class label is not available. Secondly, the feature space
trained using reconstruction loss may not be optimal for solving regression.
Thus, we propose a feature space trained using an unsupervised type
of deep learning such as an autoencoder. A recently proposed ABC clustering
loss for the autoencoder seek to minimize the difference between the
distribution of the samples by the encoder and the partitioning by
Kmeans in the autoencoder latent space. Thus, we propose the use of
ABC to train the autoencoder for regression modeling. We demonstrated
our proposed deep learning approach mainly on the PHM Challenge 2016
and we were able to outperform several of our baseline approaches
such as statistical moments, PCA and raw input. When compared to state-of-the-arts
on PHM, we were able to outperform techniques such as Ensemble approach,
Random Forest. To the best of our knowledge, mainstream deep learning
is relatively new for CMP process.

 \bibliographystyle{IEEEtran}
\bibliography{myphm.bib,allmyref.bib}

\end{document}